\documentclass[10pt,letterpaper]{article}
\usepackage{cogsci}
\usepackage{array}
\usepackage[english]{babel}
\usepackage[utf8x]{inputenc}
\usepackage[colorinlistoftodos]{todonotes}
\usepackage{tabularx}
\usepackage[natbibapa]{apacite}
\usepackage{amsmath,amssymb,amsfonts,amsthm,bm}
\usepackage{graphicx}
\usepackage{pslatex}
\usepackage{caption}
\usepackage{subcaption}

\title{Modeling Human Categorization of Natural Images \\ Using Deep Feature Representations}
  
\author{
{\large \bf Ruairidh M. Battleday (battleday@berkeley.edu)*}\\
Helen Wills Neuroscience Institute, University of California, Berkeley\\
{\large \bf Joshua C. Peterson (peterson.c.joshua@gmail.com)}\\
{\large \bf Thomas L. Griffiths (tom\_griffiths@berkeley.edu)} \\
Department of Psychology, University of California, Berkeley\\
{}\\
{\bf *} Corresponding author}

\begin{document}
\maketitle
\begin{abstract}
Over the last few decades, psychologists have developed sophisticated formal models of human categorization using simple artificial stimuli. In this paper, we use modern machine learning methods to extend this work into the realm of naturalistic stimuli, enabling human categorization to be studied over the complex visual domain in which it evolved and developed. We show that representations derived from a convolutional neural network can be used to model behavior over a database of $>$300,000 human natural image classifications, and find that a group of models based on these representations perform well, near the reliability of human judgments. Interestingly, this group includes both exemplar {\it and} prototype models, contrasting with the dominance of exemplar models in previous work. We are able to improve the performance of the remaining models by preprocessing neural network representations to more closely capture human similarity judgments.\\

\end{abstract}

\section{Introduction}
The problem of categorization---how an intelligent agent should group stimuli into discrete concepts---is an intriguing and valuable target for psychological research: it extends many influential themes in Western classical thought (see \citeauthor{AristotleCats}, \citeyear{AristotleCats}), has clear interpretations at multiple levels of analysis \citep{marr82}, and is likely fundamental to understanding human minds and advancing artificial ones \citep{CohenHandbook}. Previous categorization research has had many successes---in particular, the development of high-precision statistical models of human behavior. In this literature, human categorization data has often been accounted for with respect to either category summaries or abstractions (``prototype'' models) or stored examples in memory (``exemplar'' models) \citep{maddox1993comparing, reed72, mckinley1995investigations}. These seemingly disparate models can be unified mathematically as strategies for density estimation (parameteric and nonparameteric, respectively; see \citeauthor{ashbyar95}, \citeyear{ashbyar95}), an interpretation that enables interpolation between them, most notably in mixture density estimators \citep{rosseel02}. Fully extrapolating the probabilistic re-framing of categorization allows one to explain the rational choice among these estimators using Bayesian nonparametric methods, tying the complexity of the strategy to the availability of data to the learner \citep{griffiths2007unifying}. \\

\begin{figure}[!ht]
    \centering
	\includegraphics[width=1.0\linewidth,keepaspectratio]{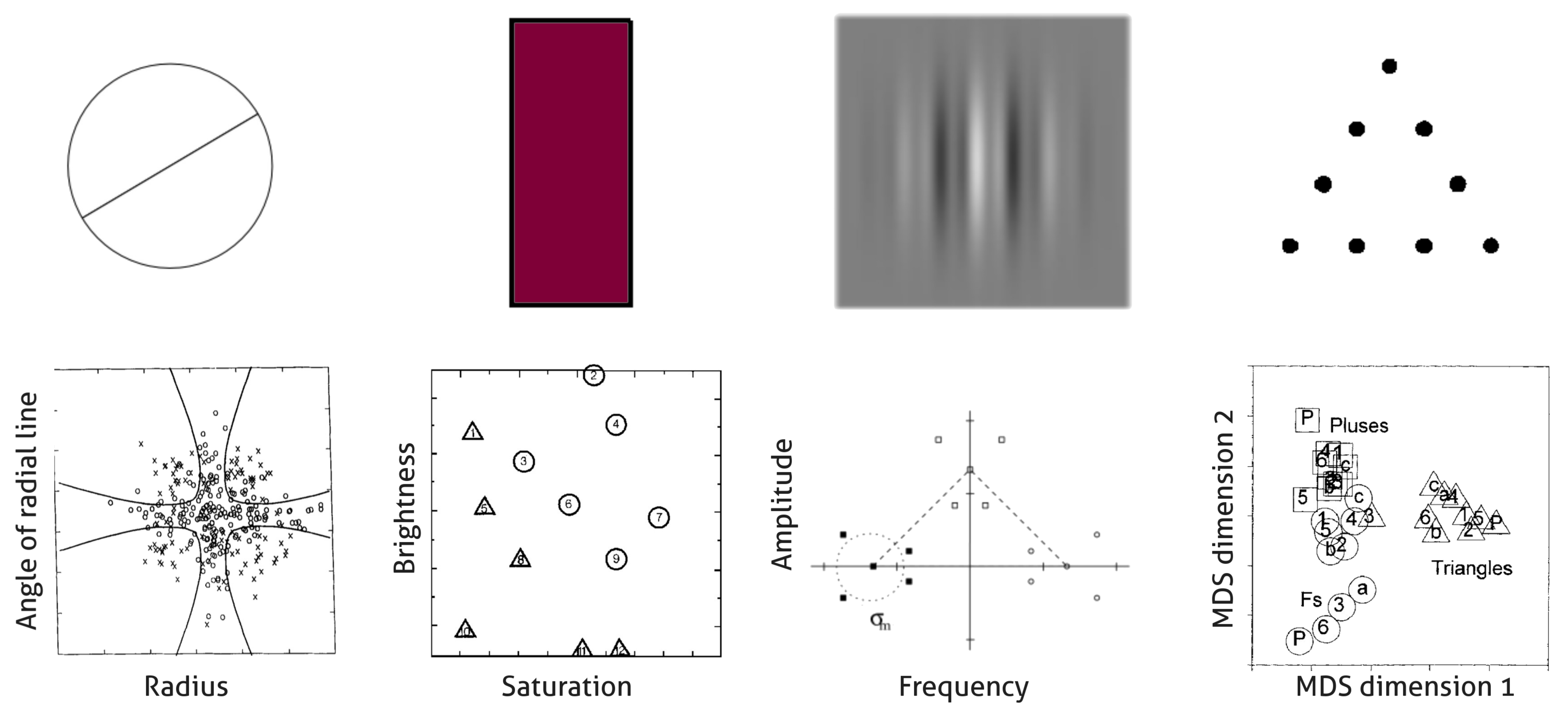}
	\caption{Stimuli from previous canonical studies of categorization. The top row shows a representative stimulus, the bottom representations of these stimuli to be used as inputs to categorization models. Modified from (left to right): \cite{mckinley1995investigations}; \cite{nosofsky1988similarity}; \cite{juettner2000scale}; \cite{palmeri2001central}.}
	\label{fig:stims}
\vspace{-3mm}
\end{figure}

While this work has been insightful and theoretically productive, we know little about how it relates to the complex visual world it was meant to describe: it derives almost exclusively from laboratory experiments using highly-controlled and simplified perceptual stimuli (Figure \ref{fig:stims}, top row), represented mathematically by hand-coded descriptions of obvious features or multidimensional-scaling (MDS) solutions of similarity judgments (Figure \ref{fig:stims}, bottom row). Human categorization abilities, by contrast, emerge from contact with the natural world, and the problems it poses. As the category divisions that result may be best understood in this context, a central challenge is to extend existing theory to account for behavior over such domains. Recent work has begun to take up this challenge \citep{nosofsky2017learning}; however, a fundamental problem remains finding appropriate psychological representations to do so for large numbers of varied naturalistic stimuli. \\

Developments in machine learning suggest one means to solve this problem. Tackling natural-image classification from an engineering perspective, computer scientists have achieved human-level accuracy using deep neural networks loosely inspired by the structure of the human brain. These networks learn representations that are used to optimize classification of large sets of natural images \citep{lecun2015deep}, and hence provide a source of representations of complex naturalistic stimulus structure that can be used as input to psychological models of categorization. While it is unclear whether such models resemble human categorization or feature learning, the representations they learn are nevertheless apparently relevant to information that humans use to judge stimulus similarity, and have been shown to provide a reasonable basis for approximating human representations in psychological experiments \citep{lake2015deep, peterson2016adapting}. \\

\begin{figure}[hb!]
\centering
\includegraphics[width=1\linewidth,keepaspectratio]{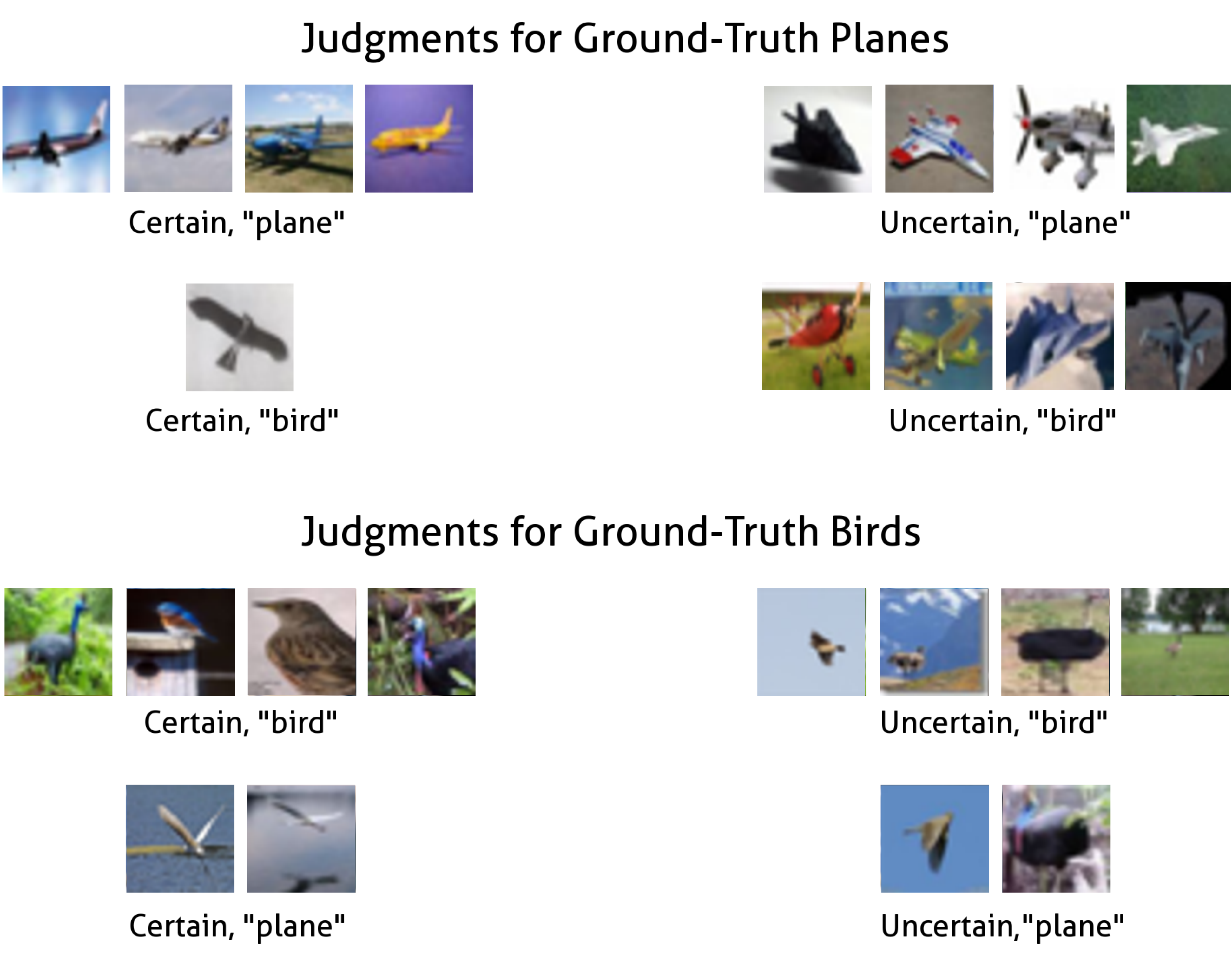}
\vspace{-2mm}
\caption{Consensus human judgments for ground-truth images of planes and birds. Sets of images vary in the proportion of subjects that categorize them as planes or birds.}
\label{fig:judgments}
\vspace{-5mm}
\end{figure}

In this study, we show that the representations extracted from a convolutional neural network (CNN) can be used as input for formal prototype and exemplar models of categorization, paving the way towards a wider and more nuanced exploration of human behavior. Moreover, we do so outside of the traditional laboratory setting, using representations from three layers of a canonical CNN to model human behavior over a massive dataset of crowd-sourced natural-image category judgments (see Figure \ref{fig:judgments}). We find that models based on CNN representations perform well, close to the reliability of human judgments. Surprisingly, although an exemplar model performs best overall, several variants of the prototype model are nearly as accurate, a finding that contrasts with what might be expected based on previous research, and one that highlights the importance of the representational space on the relative performance of categorization models. Models making more rigid assumptions about category structure, based on CNN representations alone, perform less well. However, we show that over some layers of the neural network their performance can be improved by integrating salient information about human behavior in an alternate way: pre-transforming CNN representations to more closely approximate human ones. These results demonstrate a promising route by which modern machine learning methods can be developed as a tool to extend traditional cognitive modeling of categorization into more representative domains. \\

\subsection{Modeling categorization of natural images}
We begin with a brief review of categorization models and convolutional neural networks. More mathematical details may be found in the Methods section.
\subsubsection{Categorization models}
It seems intuitive that we categorize a novel stimulus based on its similarity to previously-learned concepts and categories. This motivates the comparison of formal models within a common framework: categorization as the assignment of a novel stimulus, $\mathbf{y}$, to a category, $C$, based on some measure of similarity, $S(\mathbf{y},t)$, between the feature vectors of $\mathbf{y}$ and those of existing category members (expressed in a summary statistic, $t_C = f({\mathbf{x}: \mathbf{x} \in C})$). We may now specify a model by a summary statistic, $t$, the similarity computation, $S$, and a function that links similarity scores for each category to the probability of selecting that category. \\

The summary statistic, $t$, represents the properties of categories that are necessary inputs for the similarity calculation under different strategies. In the psychological literature, two canonical strategies have been developed regarding these properties. In a {\bf prototype model} (for example, \citeauthor{reed72}, \citeyear{reed72}), a category prototype---the average of category members---is used for comparison: $t_C$ becomes the central tendency of the members of category $C$, $\mu_C$.
In an {\bf exemplar model} (for example, \citeauthor{nosofsky86}, \citeyear{nosofsky86}), all existing category members are used. Accordingly, $t_C$ represents all existing members or ``exemplars'' of category $C$. 
For the similarity calculation, we follow Shepard (\citeyear{shepard87}) and use an exponentially-decreasing function to relate distance in stimulus feature space to similarity. We also take $S$ to be an additive function: if $t$ is a vector, $S$ becomes the summation of the similarities between $\mathbf{y}$ and each element of $t$. We then use the Luce-Shepard choice rule \citep{luce59, shepard1957stimulus} to determine the likelihood of a single categorization, made over our two categories (plane ($P$) and bird ($B$)):
\begin{equation}
p(\mbox{Guess\, } ``Plane\,"|\mathbf{y}) \,  = \, \frac{S(\mathbf{y},\, t_P)^{\gamma}}{\, S(\mathbf{y},\, t_P)^{\gamma} \, + \, S(\mathbf{y}, \, t_B)^{\gamma}}.
\label{eq:LS_basic}
\end{equation}

\subsubsection{Convolutional Neural Networks}
\begin{figure*}[!h]
    \centering
	\includegraphics[width=0.8\linewidth,keepaspectratio]{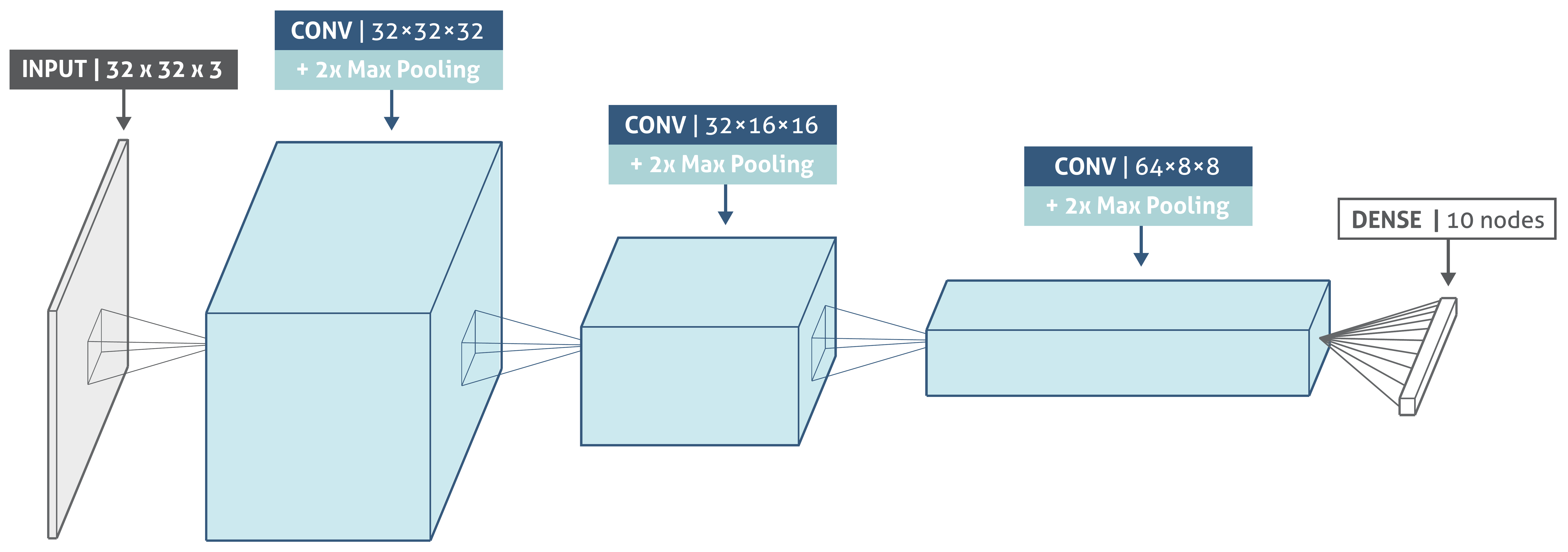}
	\caption{The CNN architecture we used to train on the \textsc{CIFAR10} image dataset to supply three sets of representation, one from each layer. Pixel-valued RGB inputs for each image are fed into a series of stacked non-linear transformations, each of which is composed of a linear transformation, and a set of non-linearities. The final ``softmax'' layer is equivalent to a simple multivariate-Gaussian classifier.}
	\label{fig:AlexNet}
\vspace{-3mm}
\end{figure*}

Deep CNNs provide rich, transferable representations of natural images that enable state-of-the-art performance on many core problems in machine vision \citep{lecun2015deep}. They pass pixel-level input data through a series of processing layers, which either apply a convolutional filter to the activation of nodes in the previous layer, or pool a subset of them. Node activations at each layer form a vector representation of the image that is increasingly abstract, and eventually input to a simple parameteric classifier (see Figure \ref{fig:AlexNet} for a schematic of the CNN we use in this paper). Representation vectors from any layer can then be directly input into the categorization models described above. Beyond their use for flat-object classification, these representations have been shown to best predict brain activity in visual cortices \citep{agrawal2014pixels,mur2013human} and human similarity judgments for natural images \citep{peterson2016adapting}. \\

\section{Study 1: Fitting categorization models to human behavior using CNN representations}
\subsection{Design}
In our first study, we fit several variants of traditional prototype and exemplar categorization models to a large dataset of human categorization decisions over natural images, using stimulus representations from multiple layers of our CNN.
\subsection{Methods}
\subsubsection{Stimuli}
The human decisions and accompanying CNN representations we investigate are based on the \textsc{CIFAR10} dataset \citep{krizhevsky2009cifar}, which comprises 60,000  $32\times32$ color images from 10 categories of natural objects. Human judgments were collected for a subset of two categories: birds ($1,005$ images) and planes ($1,032$ images). The particular images were chosen based on uncertainty sampling: a method of increasing sample value by using intermediate models to present the stimuli they were least certain about to participants (for details, see \citeauthor{Haas15}, \citeyear{Haas15}). 
\subsubsection{Human behavioral data}
Our behavioral dataset consists of $302,778$ human categorization decisions made over this stimulus set---to our knowledge, the largest reported in a single study to date. Participants saw an image, and were asked whether it was a bird or a plane. These data were originally collected as part of a large project to improve crowd-sourcing latency, and have not been explored in a psychological context \citep{Haas15}. The mean number of judgments per image was $149$ (range: $106 \, - \, 201$).

\subsubsection{Deep representations}
\begin{figure*}[!h]
 \centering
 \includegraphics[trim={0mm 0mm 0mm 0mm   },clip,width=\linewidth,keepaspectratio]{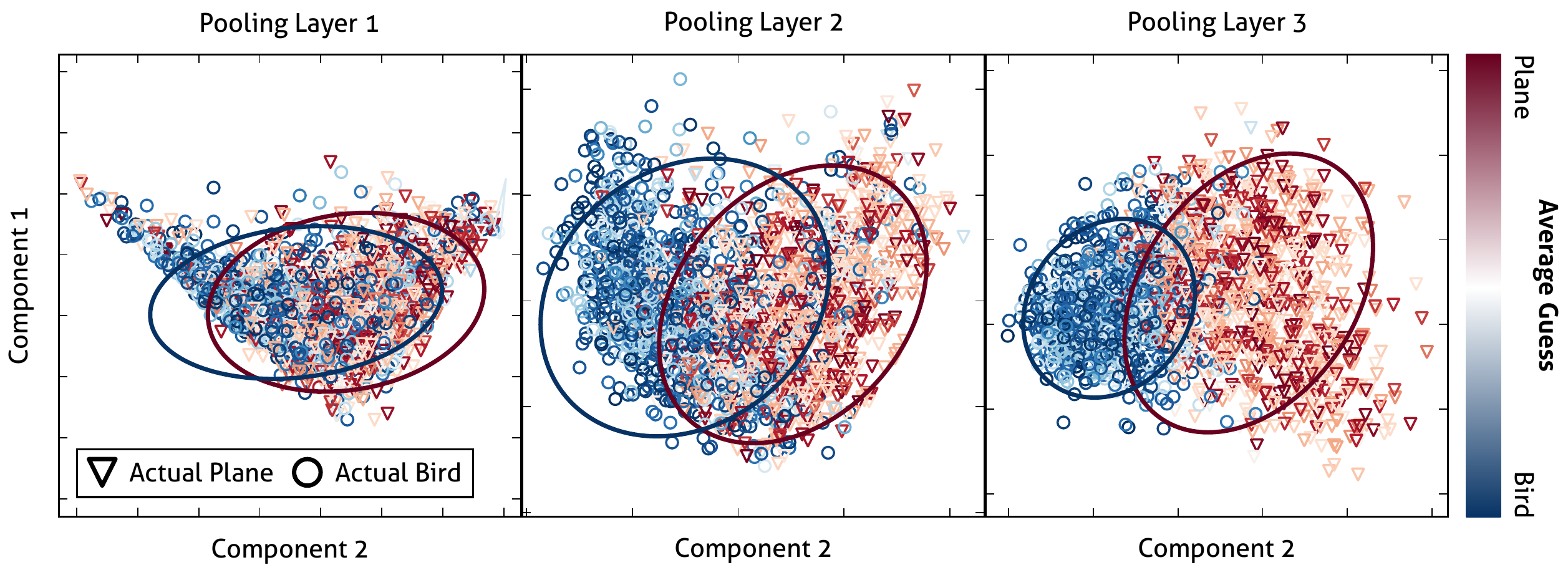}
 \caption{Two-dimensional principal component projections of the three layers of our network, colored by average guess. 95\% covariance ellipses overlaid, colored by category.}
 \label{fig:PCA_plots}
\end{figure*}

We extract feature representations for each of our stimuli from all three major layers of a simplified version of the popular AlexNet CNN \citep{krizhevsky2012imagenet}, pre-trained on the \textsc{CIFAR10} dataset to an overall 10-class classification accuracy of 82\% using Caffe \citep{jia2014caffe}; this network is depicted in Figure \ref{fig:AlexNet}. Two-dimensional principal component projections of these representations are shown in Figure \ref{fig:PCA_plots}, colored according to human judgments for the corresponding images. We use this network because it has a simple architecture that allows for easier exploration of layers while maintaining classification accuracy in the ballpark of much larger, state-of-the-art variants.

\subsubsection{Categorization models}
We can reduce the likelihood of each judgment, given in Equation \ref{eq:LS_basic}, as follows:
\begin{equation}
p(\mbox{Guess\, } ``Plane\,"|\mathbf{y}) \, = \, \frac{1}{1 + e^{ \, \gamma \, \log \left (\frac{S(\mathbf{y}, \, t_B)}{S(\mathbf{y} ,\, t_P)}\right )} }.
\end{equation}
This defines a sigmoid function around the classification boundary, in which $\gamma$ is a freely-estimated response-scaling parameter that controls its slope, and therefore degree of determinism. As $\gamma\to\infty$ it becomes deterministic, and as $\gamma\to 0$ it reduces to random responding. When formulated in this manner, the prototype model is equivalent to a multivariate-Gaussian classifier, and the exemplar model to {\it k}-nearest-neighbors classifier with distance weighting. \\

To evaluate the predictions of these models against human data, we record the category label, $c_i$, that the participant gives to the stimulus ${\bf y}_i$. We then compute the log-likelihood of the $N$ human guesses under the model:
\begin{equation}
\mathcal{L} \, = \, \sum_{i=1}^{N} \, \log \, \frac {1} {1 + e^  { -\, \gamma \, c_i \, \log \left (\frac    {S(\mathbf{y},t_B)}{S(\mathbf{y},t_P)} \right )   } },
\end{equation}
where $c_i$ is the label participant gives to the stimulus ${\bf y}_i$, and takes the value $-1$ for $P$, and $1$ for $B$, acting to invert the difference of distances appropriately. Prototype and exemplar models differ in how their similarity to a category, $S(\, {\bf y} ,\, t_C)$, is calculated.  \\

\subsubsection{Prototype model variants}
For prototype models, similarity to a category is taken to be a exponential function of the negative squared Mahalanobis distance between a stimulus vector ${\bf y}$ and the category prototype:
\begin{equation}
S(\, {\bf y} ,\, t_C) \, = \,e^{\, -md_C(\mathbf{y})},
\label{eq:pt_sim}
\end{equation}
leading to the following general log-likelihood for a prototype model:
\begin{align}
 \mathcal{L}_{P} \, =  \, \sum_{i=1}^{N} \, \log \, \Bigg[ \frac{1}{\, 1 + \, e^{- \, \gamma \, c_i\, (md_B\, \, -  \,\,md_P)}} \Bigg] .
\end{align}

The Mahalanobis distance itself is given by the following equation:
\begin{equation}
md_C(\mathbf{y}) = 
(\mathbf{y}-\mu_C)^t \, \bm{\Sigma}_C^{-1} \, (\mathbf{y}-\mu_C),
\label{eq:mh_distance}
\end{equation}

where $\mu_C$ and $\bm{\Sigma}_C$ are the mean---or, prototype---and covariance matrix of category $C$. We can define a number of prototype models by using different strategies to estimate these two parameters for each ground-truth image category, resulting in five {\it linear} and four {\it quadratic} prototype models (see Table \ref{tab:prototypes}).\\

If $\bm{\Sigma}_C$ is the same for all categories, then the boundary at which a stimulus changes which prototype it is closest to is a hyperplane, resulting in a linear model \citep{dudahs00}. Taking the mean of category representations as prototype $\mathbf{\mu}_C$, we test the following models, which define different linear decision boundaries in feature space:

\begin{itemize}
\setlength\itemsep{1mm}
 \item Identity: $\bm{\Sigma}_C$ is the identity matrix, $\mathbf{I}$, for both cases. In this case, the Mahalanobis distance reduces to the (squared) Euclidean distance;
 \item Common Variance: $\bm{\Sigma}_C$ is a diagonal matrix, with the empirically-estimated variance of all vectors (both plane and bird) as its diagonal---i.e., $\bm{\Sigma}_P, \, \bm{\Sigma}_B \, = \, \bm{\Sigma}_{P+B}$;
 \item Common Vector Variance: $\bm{\Sigma}_C \, = \, \mathbf{c} \, \mathbf{I}$, where $\mathbf{c}$ is a vector fitted on training set data for both categories $P$ and $B$.
\end{itemize}

The first of these models is the simplest, and is equivalent to most prototype models employed throughout the literature. We may also reduce the above equation as follows, allowing us to posit two additional ``Hyperplane'' models:
\begin{equation}
\mathcal{L}_{PH} \,=\, \sum_{i=1}^N \, \log \Bigg[ \frac{1}
	{\, 1 + \, e^
	{-\, \gamma \,c_i \, (2\mathbf{y}_i^T\mathbf{v}+d)
	}}\Bigg].
\end{equation}
Here, $\mathbf{v}$ defines a $(d-1)$-dimensional decision hyperplane parallel to the midpoint of the line linking the prototypes, offset by a bias term representing the difference in squared length of the means, $d$. This method corresponds to dropping the estimation of prototypes from category representations, and instead learning the projection of the line connecting the prototypes in human consensus space into the CNN representational space; equally, it can be thought of as learning a linear transformation of CNN representational space based on behavioral data. \\

If $\bm{\Sigma}_C$ is allowed to vary across categories, then this classification boundary can take more complex non-linear forms \citep{dudahs00}. Taking the mean of category representations as prototype $\mathbf{\mu}_C$, we test the following models, which define different quadratic decision boundaries in feature space:
\begin{itemize}
\setlength\itemsep{1mm}
    \item Category Pooled Variance: $\bm{\sigma}_{C}$ is the empirically-estimated scalar-valued mean of category $C$'s variance terms. This is also known as ``poole'' or ``spherical'' variance;
    \item Category Variance: $\bm{\Sigma}_{C} $ is a diagonal matrix with the empirically-estimated variance of category $C$'s vectors as its terms;
    \item Category Scalar Variance: $c$ is a scalar fitted on the training-set data for category $C$;
    \item Category Vector Variance: $\mathbf{c} $ is a vector fitted on training set data for category $C$.
\end{itemize}

\begin{table}
\begin{center}
\caption{Prototype model variants: descriptions \& notation}
\centering
\begin{tabular}{ |l||l|l|l|}
 \hline
 \textbf{Model Name} & $\bm{\Sigma}_{p}$ & $\bm{\Sigma}_b$ & $\mathbf{N_P}$ \\
 \hline
 Identity & $\mathbf{I}$ & $\mathbf{I}$& 1\\
 Common Variance   & $\bm{\Sigma}_C$ & $\bm{\Sigma}_C$ & 1\\
 Vector Common Variance & $ \mathbf{c}\mathbf{I} $ & $ \mathbf{c}\mathbf{I} $ & 1 + $N_f$\\
 Hyperplane (no bias) & $\propto \, \bf I$ & $\propto \, \mathbf{I}$ & 1 + $N_f$ \\
 Hyperplane (bias) & $\propto \, \bf I$ & $\propto \, \mathbf{I}$ & 2 + $N_f$\\
 Category Pooled Variance &  $ \mathbf{\sigma}_{P} \, \, \mathbf{I}
 $ &$ \bm{\sigma}_{B} \, \, \mathbf{I} $ & 1\\
 Category Variance &   $\bm{\Sigma}_{P}$ & $\bm{\Sigma}_{B}$ & 1\\
 Category Scalar Variance &  $p\mathbf{I}$ & $b\mathbf{I}$ & $3$\\
 Category Vector Variance &  $ \mathbf{p}\mathbf{I} $ & $ \mathbf{b}\mathbf{I}$ & 1 + $2N_f$\\
 \hline
 \noalign{\vskip 1mm}
  \multicolumn{4}{l}{Note: $\mathbf{N_P}$ = number of model parameters,}\\
  \multicolumn{4}{l}{$N_f = $ number of feature dimensions.}
 \end{tabular}
 \label{tab:prototypes}
\end{center}
\end{table}

\subsubsection{Exemplar model variants}
In exemplar models, the similarity between ${\bf y}$ and $t_C$ is given by
\begin{align}
S(\mathbf{y}, t_C ) & = \, \,\sum_{x \in C_m}e^{-\beta d(y_i,x)^q}
\end{align}
where $q=2$ is a shape parameter, and $\beta$ is a ``specificity'' parameter. This results in the following general log-likelihood for an exemplar model is as follows:
\begin{equation}
 \mathcal{L}_{E} \, =  \, \,\sum_{i=1}^{N} \, \log \,\Bigg[ \frac{1}{ 1+ \left ( \frac{\,\sum_{b \in B}e^{- \beta \, \,d(y_i,b)^q} }{ \,\sum_{p \in P}e^{- \beta \, \,d(y_i,p)^q}}\displaystyle \right )^{-\,\gamma c_i} }\Bigg]\, .
\end{equation}

The distance between two vectors is given by:
\begin{equation}
d(y,x) = \big[\sum_{k=1}^d\,w_k \,  \mid \mathbf{x}_k \,-\,\mathbf{y}_k  \mid ^{r}\big]^{1/r} \, ,
\label{eq:distance}
\end{equation}
where the $w_k$'s are positive dimensional-scaling parameters called ``attentional weights'' that must sum to one, and we use $r=2$ (the Euclidean norm). These attentional weights serve to modify the importance of each dimension in each distance calculation, and we build two exemplar models based on them. In the ``no attention'' model we eliminate them from the calculation; in the ``attention'' model we learn them from the data (see Table \ref{tab:exemplars}).

\begin{table}
\begin{center}
\caption{Exemplar model variants: descriptions \& notation}
\centering
\begin{tabular}{ |l||l|l| }
 \hline
 \textbf{Model Name} & $w_k$'s & $\mathbf{N_P}$ \\
 \hline
 Attention & ${\bf w}$ & 2 + $N_f$ \\
 No attention & $\frac{1}{N_f}{\bf 1}$ & 2 \\
 \hline
 \noalign{\vskip 1mm}
  \multicolumn{3}{l}{Note: $\mathbf{N_P}$ = number of model parameters,}\\
  \multicolumn{3}{l}{$N_f = $ number of feature dimensions.}
 \end{tabular}
 \label{tab:exemplars}
\end{center}
\end{table}

\subsubsection{Optimization}
We learned all model parameters with 5-fold-cross-validation and early stopping, using the Adam variant of stochastic gradient descent \citep{kingmaADAM} and a batch size of 256 images. For each fold, we generated a log-likelihood score for the held-out validation set every 10 batches. The early-stopping point for each model was the trial index at which the average validation log-likelihood score across folds was minimized. For each model, we conducted a grid search over Adam's learning rate ({\it alpha}) hyperparameter, selecting the final model parameter set based on which gave the lowest cross-validated average-log-likelihood at the model's early-stopping point.\\

\subsubsection{Model comparison}
For each of our models, we present the following three measures of performance: log-likelihood, correlation with human response proportions, and Aikake Information Criterion \citep{akaike1998information}. As a baseline model, we use the raw output probabilities of the neural network for each image to give $S(\mathbf{y},\, t_P)$ and $S(\mathbf{y},\, t_B)$, normalized to sum to one for each image. For each of our models, including the CNN softmax baseline, we computed final log-likelihood scores by generating predictions for all images in our stimulus set using the averaged cross-validated parameters taken at the early-stopping point described above.\\

As an `ideal' model, we use split-half reliability, applying the Spearman-Brown correction \citep{spearman1910correlation, brown1910some}, which gives an indication of the inter-participant consistency and a ceiling on model performance. To do this, we generate 100 random half-splits of the human judgments, where each half-split contains half the human guesses for each image. For each split, we compute the correlation between the two halves and take the mean of all 100 of these correlations to get a final reliability estimate, then applying the Spearman-Brown correction \citep{spearman1910correlation, brown1910some}. To compare our categorization models to this ideal model, we again take 100 random splits of the data and compute the correlation between the average of each half and our model predictions. We then average the results for predicting the two halves and average the values for all 100 splits; these values are reported as ``correlation''.\\

We also use the Aikake Information Criterion to compare models, as, under limiting assumptions, it gives a score for each model that takes into account the relationship between the number of parameters they employ, and the log-likelihood scores they produce \citep{akaike1998information}:
 \begin{equation}
  AIC \, = \, 2k \, - \, 2\ln(\hat{\mathbf{L}}) 
 \end{equation}
where $k$ = the number of parameters in the model, and $\hat{\mathbf{L}}$ is the maximum log likelihood. (Equivalently, it can be thought of as estimating the relative information lost by using that model to represent the true underlying generative process.) \\ 

\subsection{Results}
\begin{table}[!ht]
\begin{center}
\caption{Scores for baseline and ceiling models evaluated on entire categorization dataset}
{\small
\begin{tabular*}{\linewidth}{@{\extracolsep{\fill}}lllll}
\hline
 \noalign{\vskip 1mm}
 \textbf{Model} & \textbf{LL} & \textbf{AIC} &\textbf{Correlation} & $\mathbf{N_P}$ \\
\hline
 \noalign{\vskip 1mm}
 \multicolumn{5}{l}{\textit{Baseline}} \\
 \noalign{\vskip 1mm}
 \hspace{2mm}NN Softmax & -168,152 & 336,306 & 0.67 & 1 \\
 \hline
\noalign{\vskip 1mm}
 \multicolumn{5}{l}{\textit{Ceiling}} \\
\noalign{\vskip 1mm}
 \hspace{2mm}Split-half reliability & - & - & 0.77 & - \\
 \hline
 \noalign{\vskip 1mm}
 \multicolumn{5}{l}{Note: NN = neural network, LL = log-likelihood,} \\
 \multicolumn{5}{l}{AIC = Aikake Information Criterion, $\mathbf{N_P}$ = number of parameters.}
 \end{tabular*}
}
 \label{tab:results_baseline}
\end{center}
\vspace{-3mm}
\end{table}

\subsubsection{Baseline and ceiling measures}
Our ceiling and baseline measures are shown in Table \ref{tab:results_baseline}. At $r\,=\,0.77$, our split-half reliability indicates a large amount of inter-subject variability in image classification---beyond, perhaps, what would be expected from laboratory experiments, but understandable given the complex nature and small size of images, and the inherently-decreased precision from crowdsourcing data and using uncertainty sampling to select stimuli (see \citeauthor{Haas15}, \citeyear{Haas15}). \\

We also consider the CNN softmax output as a baseline model. For each image, the softmax function takes the inner product between a matrix of learned weights and the rasterized output of the final pooling layer, and returns a probability distribution over all of the \textsc{CIFAR10} classes. These weights are learned based on minimizing classification loss over the whole \textsc{CIFAR10} dataset, which comprises 50,000 training images over 10 categories. Thus, the CNN softmax is a generous ``baseline'', as it includes many extra parameters learned over a much larger dataset for the related task of ground-truth image categorization. Consistent with this, it achieves a low log-likelihood score and high correlation. \\

\begin{figure*}[!ht]
    \centering
	\includegraphics[trim={0mm 0mm 0mm 0mm },clip,width=\linewidth,keepaspectratio]{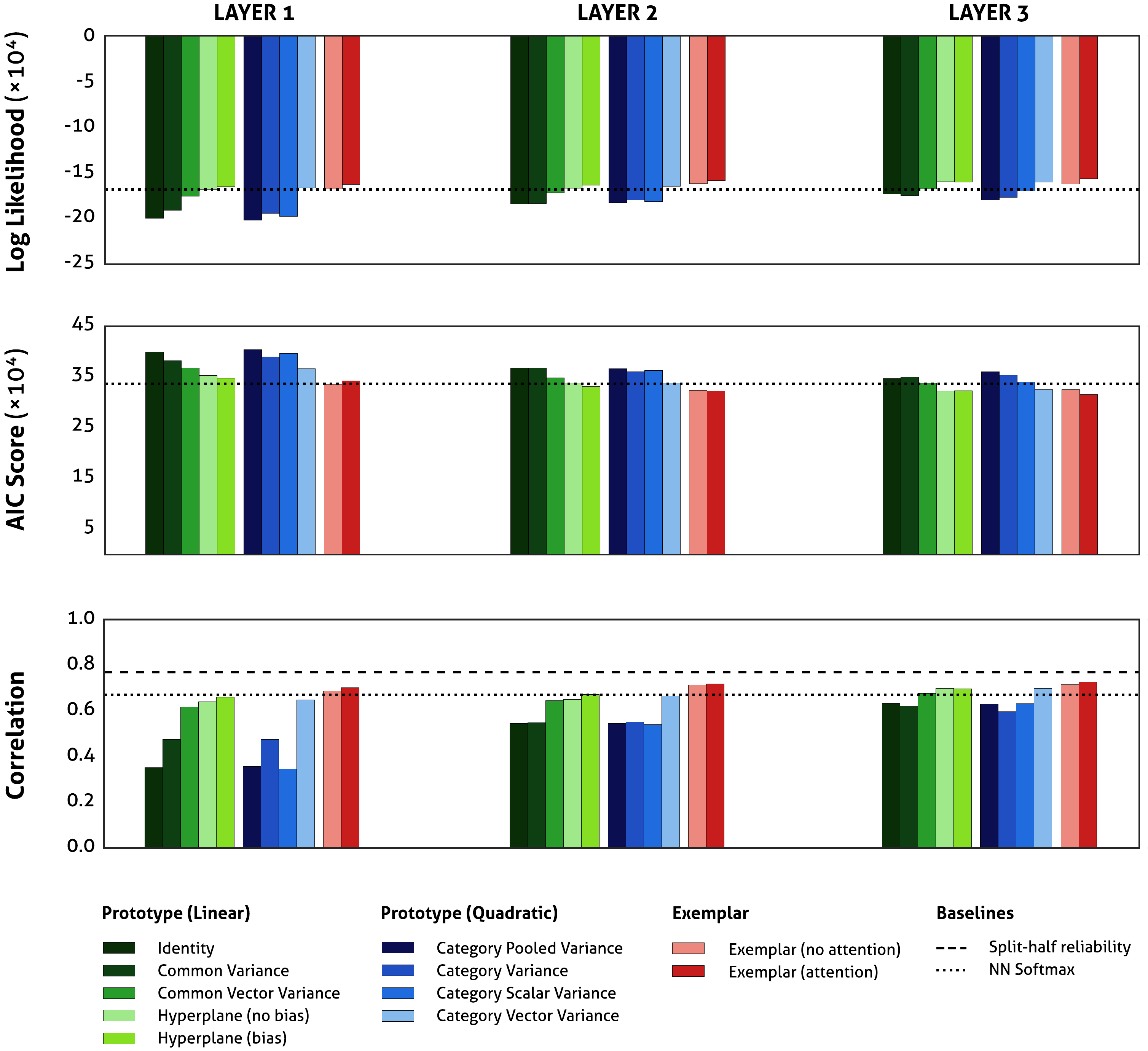}
	\caption{Model results using untransformed representations from all three layers of our convolutional neural network image classifier, reported as log likelihood (top panel), AIC score (middle panel), and correlation with human response proportions (bottom panel).}
	\label{fig:untrans_results}
\vspace{-5mm}
\end{figure*}

\subsubsection{Categorization models}
Categorization model performance using the untransformed CNN representations is shown in Figure \ref{fig:untrans_results}, with numerical scores in the Appendix. In terms of log-likelihood, five models consistently outperform the CNN baseline: the hyperplane models from the linear prototype class, with and without a bias term, the category vector variance model from the quadratic prototype class, and the exemplar models, with and without attentional weights. In general, these models have more parameters, meaning they are able to alter CNN representations using the human behavioral data. Although the exemplar model with attentional weights performs best overall, it is striking that the simple decision bounds formed by these prototype models allow them to perform nearly as well, and over all three layers. Indeed, when reviewing correlation scores, we can see that these five models are all performing close to the ceiling provided by the split-half reliability. Prototype models with fewer parameters, however, performed less well, and consistently below the baseline. These models do not incorporate information about human behavior to alter the shape of their decision boundary during training, instead estimating this from the CNN representations alone. \\

All models perform better using more abstract and lower-dimensional representations from higher layers, with the exception of the exemplar model without attentional weights. Again, this difference is largest in the prototype models with fewer parameters. While this result may have cognitive implications, it can be explained in machine learning terms. During training, the CNN has learned a feature representation that allow it to disambiguate stimuli easily in the deepest layer (Layer 3), meaning features associated with different categories become increasingly well-separated with depth. The greater degree of category overlap in more superficial layers therefore penalizes models that directly estimate categorical structure from CNN representations. In addition, with an increase in feature dimensions the generalization of solutions found during training is likely to worsen, as the ratio of stimuli to dimensions, and therefore available information to constrain solutions, decreases from around $1:1$ in Layer 3 to around $1:8$ in Layer 1. \\


In order to offer an alternate evaluation of models that takes this risk of overfitting into account, we also present model AIC scores, which penalize more highly-parameterized models. This analysis does not affect model rank in the deepest two layers, but does indicate that in Layer 1 the risk of over-parameterization outweighs the benefit in log-likelihood for all models, except the exemplar without attentional weights, compared to the CNN softmax.

\section{Study 2: Fitting models using transformed CNN representations}
\subsection{Design}
Recent work demonstrates that human similarity judgments can be used to transform vector representations of stimuli to more closely correspond to human ones \citep{peterson2016adapting}. The core strategy of these techniques is to use a learned transformation of the underlying space to increase the correlation between the human similarity scores of stimuli and some measure of vector similarity---for example, the inner product. There are two reasons for doing so: first, this approach extracts and amplifies information in stimulus representations that is relevant to the human behavior being modeled, resulting in a more faithful and interpretable conceptual structure. Second, in previous categorization research, low-dimensional MDS solutions have themselves been used as representations of more complex stimuli (for example,  \cite{palmeri2001central}). Using transformation techniques complements and extends this approach by retaining the information content of higher-dimensional spaces, but doing so in a way that directly improves the quality of conceptual structure in lower-dimensional MDS projections. In our second set of analyses, we first transform CNN representations to more closely approximate human similarity judgments and then re-evaluate our categorization models using these improved approximations.

\subsection{Method}
\subsubsection{Stimuli}
We collected similarity judgments for a randomly-selected subset of 60 birds and 60 planes from the \textsc{CIFAR10}-based categorization stimuli.
\subsubsection{Behavioral data}
We collected 10 similarity judgments between each of the $7,140$ unique pairs of these stimuli on Amazon Mechanical Turk, giving a total of $71,400$ ratings from $209$ different
participants. Participants were instructed to rate the similarity of four pairs of bird and plane images on a scale from 0 (not similar at all) to 10 (very similar). We paid workers \$0.02 per set of four comparisons. Before each task, eight example pairs were
shown to help prevent bias in early judgments. Amazon workers could repeat the task with new pairs as many times as they wanted. The result was a $120 \times 120$ similarity matrix after averaging over judgments.
\subsubsection{Transforming CNN representations}
To transform our representations, we follow the method introduced in \citeauthor{peterson2016adapting}, (\citeyear{peterson2016adapting}), which uses $L2$-regularized linear regression to increase the correlation between vector inner products and the average human similarity judgment between their corresponding images. A similarity matrix {\bf S} can be expressed as the matrix product of a feature matrix {\bf F} and a diagonal weight matrix {\bf W} \citep{shepard1979additive}, 
\begin{equation}
    {\bf S} = {\bf F}{\bf W} {\bf F}^T
\end{equation}
Given an existing feature matrix, the diagonal of ${\bf W}$ can be obtained using through ordinary least squares. This is more evident when expressing the entries of the S matrix algebraically:
\begin{equation}
  \label{eq:u}
  \begin{gathered}
\displaystyle s_{ij} = \sum_{i=1}^{N_{f}} w_{k}f_{ik}f_{jk} .
 \end{gathered}
 \end{equation}
In the context of our categorization models, we require that these weights reflect a linear transformation of squared {\it distances}; therefore, they can be further constrained to be non-negative. We use the non-negative least squares algorithm from the scipy python module, enforcing $L2$ regularization by augmenting the row space of the matrix with $d$ orthogonal vectors, whose length is controlled via the ridge parameter (in other words, by manually implementing ridge regression---see \citeauthor{vanWRidge}, \citeyear{vanWRidge}). We find the optimal regularization parameter using a grid search over values with 5-fold cross-validation. We then retrain the model with these parameters on the whole dataset to yield the final diagonal coefficient matrix, ${\bf W}$. We use this transformation to generate a second set of representations for our images by pre-multiplying the representation matrix with the element-wise square root of the weight matrix, which is equivalent to the calculation described above.

\subsection{Results}
\subsubsection{Transforming representations}
\begin{figure*}[t!]
\centering
 \includegraphics[width=0.8\linewidth]{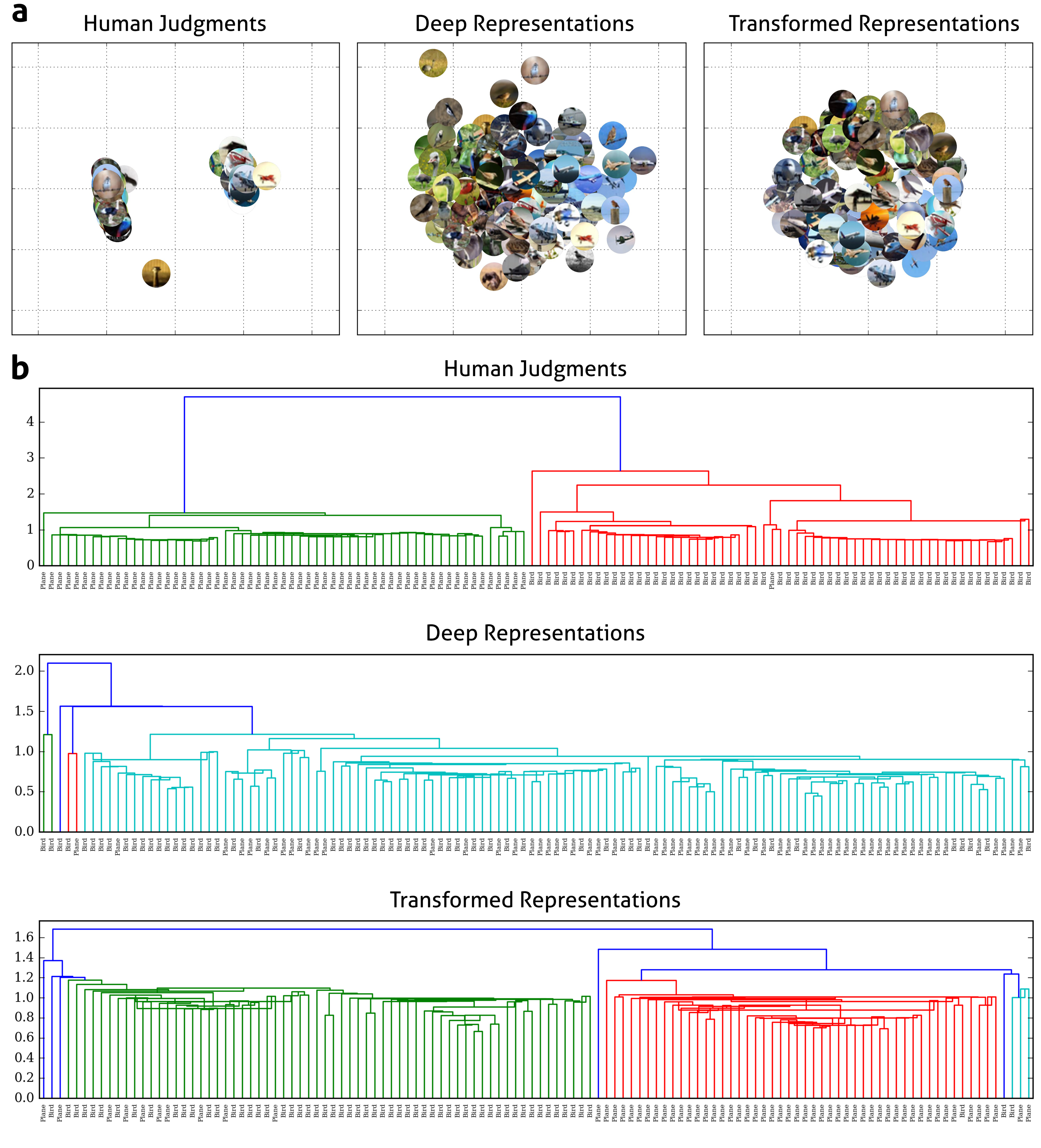}
\caption{MDS and dendrogram for similarity data, untransformed, and transformed representations for representative layer (Layer 2). a) MDS solutions for similarity stimuli; b) Dendrogram of similarities. Lines colored by cluster, where cluster membership is determined by distance to other entities below a threshold.}
\label{fig:judgments_L2}
\end{figure*}

Using the linear transformation described above, we are able to substantially increase their correlation to human similarity ratings, especially in the deepest layers (see Table \ref{tab:results_similarity}). In improving pair-wise correlation, this transformation also recovers key global structure in stimulus organization; see Figure \ref{fig:judgments_L2}. 

\begin{table}[!h]
\begin{center}
\caption{Correlation with human similarity judgments}
{\small
\begin{tabular*}{\linewidth}{@{\extracolsep{\fill}}llll}
\hline
 \noalign{\vskip 1mm}
 \textbf{Representation type} & \textbf{Layer 1} & \textbf{Layer 2} &\textbf{Layer 3} \\
\hline
 \noalign{\vskip 1mm}
 \noalign{\vskip 1mm}
 \hspace{2mm}Untransformed & 0.004 & 0.03 & 0.12 \\
 \hspace{2mm}Transformed  & 0.18 & 0.28 & 0.37 \\
 \hline
 \noalign{\vskip 1mm}
  \multicolumn{4}{l}{Note: Values shown are correlation ($r$-squared) between stimulus}\\
  \multicolumn{4}{l}{vector inner products and mean human similarity judgments.}\\
 \end{tabular*}
}
 \label{tab:results_similarity}
\end{center}
\vspace{-3mm}
\end{table}

\begin{figure*}[!ht]
    \centering
	\includegraphics[trim={0mm 0mm 0mm 0mm },clip,width=\linewidth,keepaspectratio]{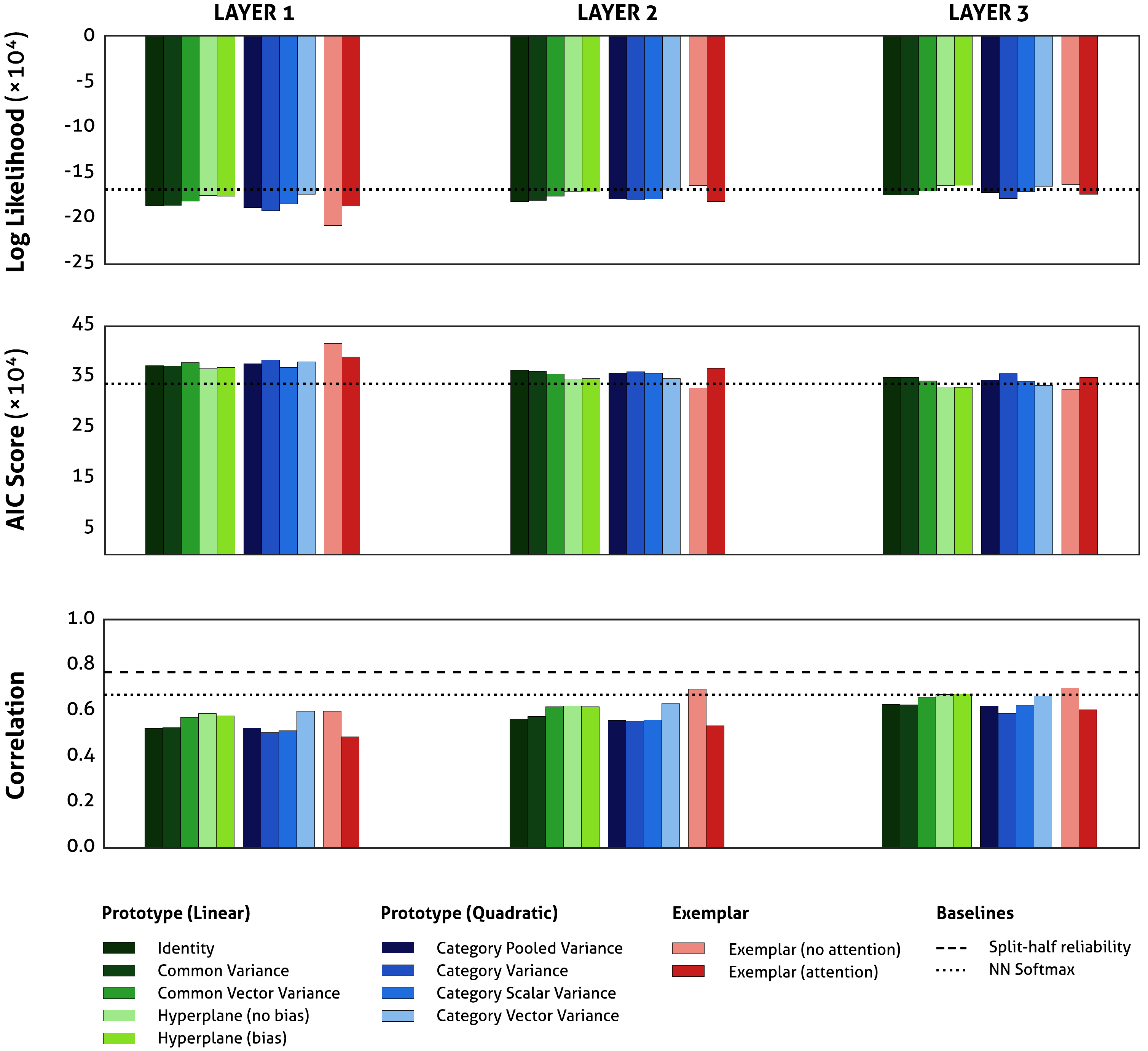}
	\caption{Model results using transformed representations from all three layers of our convolutional neural network image classifier, reported as log likelihood (top panel), AIC score (middle panel), and correlation with human response proportions (bottom panel).}
	\label{fig:trans_results}
\vspace{-5mm}
\end{figure*}

\subsubsection{Categorization models}
The results from evaluating our categorization models on these transformed CNN representations are shown in Figure \ref{fig:trans_results}, with numerical scores shown in the Appendix. For the prototype models, the same general pattern holds across layers: models that are able to augment computation on CNN representations with more information about human behavior perform better, with the same models performing best. With the exemplar models, the picture is more complex. The exemplar model without attentional weights performs well in the deepest two layers, scoring higher than baseline. The exemplar model with attentional weights, however, does not, and only performs best in the most {\it superficial} layer. \\

When comparing these results to Figure \ref{fig:untrans_results}, an interesting pattern emerges. The transformation {\it improves} the performance of prototype models with fewer parameters---which form their decision boundaries based on the CNN representations alone---in Layers 1 and 2. The Category Scalar Variance model could also be included in this class, as it only incorporates very coarse information from human behavior; and with this model the pattern also holds. The opposite effect holds with the top-performing, more highly-parameterized models, including the exemplar model with attentional weights: scores are negatively impacted by the transformation, over all layers. The most obvious explanation  for these findings is that the transformation eliminates some dimensions that are not important for capturing similarity relations, in addition to emphasizing those that are. This may have the effect of regularizing basic models, especially in higher-dimensional feature spaces, but simultaneously penalizing more highly-parameterized ones. These can then no longer exploit information in the eliminated dimensions that is nonetheless relevant to categorization. Evidence for this theory comes from inspection of the transformation weight matrices (data not shown) in conjunction with the MDS solutions of human similarity judgments and CNN inner-product spaces. For example, in Figure \ref{fig:judgments_L2} it is clear the CNN has already formed an inner-product space to optimize categorization. The transformation increases the separation of stimuli in different categories, but it does so by eliminating the majority of dimensions. 

\section{Discussion}
Our analysis goes beyond typical evaluations of categorization models in several ways. First, we use a large collection of natural images as stimuli, enabling us to study human categorization in a domain that is representative of the environment we have evolved and learned within, and theorized about.  
Second, we are able to do so because we use state-of-the-art methods from computer vision to {\it estimate} the structure of these stimuli. This contrasts with previous work, in which a small number of {\it a priori}-identified features were manipulated to {\it define} and differentiate categories, and as a consequence were limited to simple artificial stimuli. 
Finally, we offset the modelling uncertainty these advances introduce by using large crowdsourced behavioral datasets to more finely assess graded category membership over stimuli  and improve the utility of our representations for them.
Taken together, our results show that using representations derived from CNNs makes it possible to apply psychological models of categorization to complex naturalistic stimuli, and that the resultant models make competitive predictions about human behavior. This approach naturally complements and extends related work seeking to apply these models to natural images and categories, but relying on low-dimensional similarity-based or explicitly-given feature spaces \citep{nosofsky2017learning, stormsdr00}.\\

Our most general finding is that categorization models that incorporate CNN representations predict human categorizations over natural images well---in particular, those that are able to augment CNN representations with information about human categorizations through free parameters. However, we are still able to use information about human behavior to improve the performance of less-flexible models by transforming their representational substrate to more closely reflect properties of its human counterpart. This is theoretically interesting because it indicates there is enough latent information and flexibility in the ground-truth-trained CNN representations to harness for such a related task. Further, it is practically encouraging because it shows we can successfully draw on preexisting representations and behavioral datasets for our model inputs, rather than procuring them for individual experiments at heavy computational cost. Preprocessing machine learning representations in this manner is a field in its infancy, and we are likely to see further benefits as more complex transformations are taken into account, along with classical considerations about the relative timing of similarity-judgments with respect to the main task \citep{palmeri2001central}.

Working with these complex, naturalistic stimuli reveals a potentially more nuanced view of human categorization. The broad consensus from decades of laboratory studies using simple artificial stimuli was that people could learn complex category boundaries of a kind that could only be captured by an exemplar model (for example, \citeauthor{mckinley1995investigations}, \citeyear{mckinley1995investigations}). Extrapolating from these results, we might imagine that human categorization should be thought of in terms of learning complex category boundaries in simple feature-based representations. Our results outline a different perspective. The representations formed by the CNN are complex, and within those complex representational spaces, simple category boundaries seem sufficient to capture human behavior. When we think about the cases that inspire us to theorize about categorization---of children learning to categorize furry animals as cats and dogs, say---it seems plausible that this story, developed with much more realistic stimuli, might be a reasonable alternative. \\

Despite its attractiveness, there are important limitations to our analysis. One caveat comes from the source of the representations of images that we used to obtain our results: the CNN that generated them was explicitly trained to classify images into categories, including the categories of birds and planes. It does so by trying to form a representation in which a simple boundary is sufficient to pick out one category from another. In this sense, these representations should be expected to favor prototype models. While this is a worthy concern, we don't think it significantly detracts from our results. First, as illustrated in Figure \ref{fig:judgments}, people's judgments often don't agree with the ground truth that the network was trained on, so capturing human performance using these representations is non-trivial. Second, we regard the primary contribution of our results to be an \textit{existence proof} that representations exist for complex natural stimuli that allow prototype models to perform similarly to exemplar models---illustrating that complex representations and simple boundaries provide a reasonable alternative to simple representations and complex boundaries for capturing how people reason about natural categories. We simply don't have other representations of these images that lead to better performance in predicting human behavior. Given this, an important direction for future work will be obtaining representations from other state-of-the art machine learning algorithms applied to images, including from unsupervised-learning models (for example, \citeauthor{yu2016searching}, \citeyear{yu2016searching}), to evaluate the impact of classification training on our results. \\

Categorization has traditionally been regarded as distinct from feature learning. However, our findings suggest these dual processes be considered together. When thinking about humans, feature representations are likely to have been learned early on, through a slow, data-driven learning process. Given these considerations, one might expect psychological representations to reflect the natural world, such that categorization of natural stimuli is made as efficient and as simple as possible. On the other hand, artificial or unlikely stimuli may at times carve out awkward boundaries in these spaces, which perhaps underlies the success of exemplar models up to this point. Including feature-learning in the evaluation of human categorization has been called for before \citep{schynsgt98}, and developing a deeper understanding of how these processes interact is an important next step towards more fully characterizing human categorization. \\

\bibliographystyle{apacite}
\setlength{\bibleftmargin}{.125in}
\setlength{\bibindent}{-\bibleftmargin}
\bibliography{BattledayMain}
\newpage
\section{Appendix}
Below we present the numerical results for our models, using untransformed and transformed representations (Tables \ref{tab:results_un} and \ref{tab:results_trans}, respectively).\\
\bigskip
\bigskip
\bigskip
\bigskip
\vspace{1.5mm}
\begin{table}[!h]
\begin{center}
\caption{Scores for nine prototype and two exemplar models shown for all three network layers: untransformed representations}
{\small
\begin{tabular*}{\linewidth}{@{\extracolsep{\fill}}llll}
\hline
 \noalign{\vskip 1mm}
 \textbf{Model} & \textbf{Layer 1} & \textbf{Layer 2} &\textbf{Layer 3} \\
\hline
 \noalign{\vskip 1mm}
 \multicolumn{4}{l}{\textit{Prototype - Linear}} \\
 \noalign{\vskip 1mm}
 \hspace{2mm}Identity (LL) & -199,716 & -183,976 &-173,412 \\ 
 \hspace{4mm}(AIC score) & { }399,435   &{ }367,953   &{ }346,826\\ 
 \hspace{4mm}(Correlation) & { }0.35 & { }0.54 & { }0.63  \\
 \hspace{2mm}Common Variance &-191,063 &-183,701  &-174,959 \\ 
 \hspace{4mm}(AIC score) & { }382,127  & { }367,404  &{ }349,919  \\ 
 \hspace{4mm}(Correlation) & { }0.47 & { }0.55& { }0.62  \\
 \hspace{2mm}Common Vector Variance  &-175,596 &-172,123&-167,787 \\ 
 \hspace{4mm}(AIC score) & { }367,579 & { }348,346 &{ }337,626 \\ 
 \hspace{4mm}(Correlation) & { }0.62 & { }0.64 & { }0.68  \\
 \hspace{2mm}Hyperplane (no bias)  & { }-168,223 & { }-166,855  &{\bf -160,062} \\ 
 \hspace{4mm}(AIC score)  & { }352,834  & { }337,809   &{ }{\bf 322,177}  \\ 
 \hspace{4mm}(Correlation)  & { }0.64 & { }0.65  & { }{\bf 0.70}  \\
 \hspace{2mm}Hyperplane (bias) & { }{\bf -165,480} & { }{\bf -163,492}  &-160,195 \\ 
 \hspace{4mm}(AIC score)  & { }{\bf 347,349}  & { }{\bf 331,087}   &{ }322,442 \\ 
 \hspace{4mm}(Correlation)  & { }{\bf 0.66} & { }{\bf 0.67}  & { }0.70  \\
 \hline
\noalign{\vskip 1mm}
 \multicolumn{4}{l}{\textit{Prototype - Quadratic}} \\
\noalign{\vskip 1mm}
 \hspace{2mm}Category Pooled Variance &-202,026  &-182,958  &-180,102 \\ 
 \hspace{4mm}(AIC score)  & { }404,054 & { }365,917 &{ }360,206  \\ 
 \hspace{4mm}(Correlation) & { }0.36 & { }0.55  & { }0.63  \\
 \hspace{2mm}Category Variance &-194,672  &-180,047  &-176,816 \\ 
 \hspace{4mm}(AIC score)  & { }389,346 & { }360,097 &{ }353,633\\ 
 \hspace{4mm}(Correlation) & { }0.48 & { }0.55  & { }0.60  \\
 \hspace{2mm}Category Scalar Variance  &-197,974 &-181,409 &-169,942 \\ 
 \hspace{4mm}(AIC score)  & { }395,954  & { }362,823  &{ }339,889  \\ 
 \hspace{4mm}(Correlation)  & { }0.35  & { }0.54  & { }0.63  \\
 \hspace{2mm}Category Vector Variance &{\bf -166,784} &{\bf -164,728} & {\bf -160,278} \\ 
 \hspace{4mm}(AIC score) & { }{\bf 366,342} & { }{\bf 337,654} &{ }{\bf 324,658} \\ 
 \hspace{4mm}(Correlation) & { }{\bf 0.65} & { }{\bf 0.67} & { }{\bf 0.70}  \\
\hline
\noalign{\vskip 1mm}
 \multicolumn{4}{l}{\textit{Exemplar - Nonparameteric}} \\
 \noalign{\vskip 1mm}
 \hspace{2mm}Exemplar (no attention) &-167,756 &-161,890 &-162,430 \\ 
 \hspace{4mm}(AIC score) & { }{\bf 335,516} & { }323,784 &{ }324,864   \\ 
 \hspace{4mm}(Correlation) & { }0.69 & { }0.71 & { }0.71  \\
 \hspace{2mm}Exemplar (attention) & {\bf -162,942}& {\bf -158,882}& {\bf -156,442} \\ 
 \hspace{4mm}(AIC score) & { } 342,272 & { }{\bf 321,863} &{ }{\bf 314,935}  \\ 
 \hspace{4mm}(Correlation) & { }{\bf 0.70} & { }{\bf 0.72} & { }{\bf 0.73}  \\
 \hline
 \noalign{\vskip 1mm}
  \multicolumn{4}{l}{Note: LL = log-likelihood, AIC = Aikake Information Criterion.}\\
  \multicolumn{4}{l}{{\bf Bold font} indicates best in each class of models.}
 \end{tabular*}
}
 \label{tab:results_un}
\end{center}
\vspace{-3mm}
\end{table}

\begin{table}[!h]
\begin{center}
\caption{Scores for nine prototype and two exemplar models shown for all three network layers: transformed representations}
{\small
\begin{tabular*}{\linewidth}{@{\extracolsep{\fill}}llll}
\hline
 \noalign{\vskip 1mm}
 \textbf{Model} & \textbf{Layer 1} & \textbf{Layer 2} &\textbf{Layer 3} \\
\hline
 \noalign{\vskip 1mm}
 \multicolumn{4}{l}{\textit{Prototype - Linear}} \\
 \noalign{\vskip 1mm}
 \hspace{2mm}Identity (LL) & -186,265 & -181,615 &-174,405\\ 
 \hspace{4mm}(AIC score) & { }372,533   &{ }363,233   &{ }348,813 \\ 
 \hspace{4mm}(Correlation) & { }0.52 & { }0.57 & { }0.63  \\
 \hspace{2mm}Common Variance &-185,900 &-180,501  &-174,567 \\ 
 \hspace{4mm}(AIC score) & { }371,802  & { }361,005  &{ }349,136 \\ 
 \hspace{4mm}(Correlation) & { }0.53 & { }0.58& { }0.63  \\
 \hspace{2mm}Vector Common Variance  &-181,017 &-175,658&-170,049 \\ 
 \hspace{4mm}(AIC score) & { }378,423 & { }355,416 &{ }342,149  \\ 
 \hspace{4mm}(Correlation) & { }0.57 & { }0.62 & { }0.66  \\
 \hspace{2mm}Hyperplane (no bias)  & { }{\bf -174,911} & { }{\bf -170,924}  &-164,027 \\ 
 \hspace{4mm}(AIC score)  & { }{\bf 366,211}  & { }{\bf 345,948}   &{ }330,106  \\ 
 \hspace{4mm}(Correlation)  & { }{\bf 0.59} & { }{\bf 0.62}  & { }0.67  \\
 \hspace{2mm}Hyperplane (bias) & { }-175,942 & { }-171,293  &{\bf -163,717} \\ 
 \hspace{4mm}(AIC score)  & { }368,273  & { }346,687   &{ }{\bf 329,485}  \\ 
 \hspace{4mm}(Correlation)  & { }0.58 & { }0.62  & { }{\bf 0.67}  \\
 \hline
\noalign{\vskip 1mm}
 \multicolumn{4}{l}{\textit{Prototype - Quadratic}} \\
\noalign{\vskip 1mm}
 \hspace{2mm}Category Pooled Variance &-188,054  &-178,647  &-171,802 \\ 
 \hspace{4mm}(AIC score)  & { }376,110 & { }357,296 &{ }343,606  \\ 
 \hspace{4mm}(Correlation) & { }0.53 & { }0.56  & { }0.62  \\
 \hspace{2mm}Category Variance &-191,562  &-179,941  &-178,081 \\ 
 \hspace{4mm}(AIC score)  & { }383,126 & { }359,884 &{ }356,164\\ 
 \hspace{4mm}(Correlation) & { }0.50 & { }0.55  & { }0.59  \\
 \hspace{2mm}Scalar Category Variance  &-184,156 &-178,482 &-170,681 \\ 
 \hspace{4mm}(AIC score)  & {\bf 368,318}  & { }356,969  &{ }341,368  \\ 
 \hspace{4mm}(Correlation)  & { }0.51  & { }0.56  & { }0.62  \\
 \hspace{2mm}Vector Category Variance &{\bf -173,636} &{\bf -169,158} &{\bf -164,707} \\ 
 \hspace{4mm}(AIC score) & { }380,046 & { }{\bf 346,514} &{ }{\bf 333,516}  \\ 
 \hspace{4mm}(Correlation) & { }{\bf 0.60} & { }{\bf 0.63}& { }{\bf 0.67}  \\
\hline
\noalign{\vskip 1mm}
 \multicolumn{4}{l}{\textit{Exemplar - Nonparameteric}} \\
 \noalign{\vskip 1mm}
 \hspace{2mm}Exemplar &-207,919 &{\bf -164,062} &{\bf -162,628} \\ 
 \hspace{4mm}(AIC score) & { }415,841 & { }{\bf 328,128} &{ }{\bf 325,259}  \\ 
 \hspace{4mm}(Correlation) & { }{\bf 0.60} & { }{\bf 0.69} & { }{\bf 0.70}  \\
 \hspace{2mm}Exemplar (attn) & {\bf -186,544}& -181,384&-173,510 \\ 
 \hspace{4mm}(AIC score) & { } {\bf 389,477}& { }366,869 &{ }349,073  \\ 
 \hspace{4mm}(Correlation) & { }0.49 & { }0.54 & { }0.60  \\
 \hline
 \noalign{\vskip 1mm}
  \multicolumn{4}{l}{Note: LL = log-likelihood, AIC = Aikake Information Criterion.}\\
  \multicolumn{4}{l}{{\bf Bold font} indicates best in each class of models.}
 \end{tabular*}
}
 \label{tab:results_trans}
\end{center}
\vspace{-3mm}
\end{table}

\end{document}